\documentclass[10pt,twocolumn,letterpaper]{article}

\usepackage[pagenumbers]{cvpr}

\definecolor{cvprblue}{rgb}{0.21,0.49,0.74}
\usepackage[pagebackref,breaklinks,colorlinks,allcolors=cvprblue]{hyperref}

\def\confName{}
\def\confYear{}

\title{Identity-Conditioned Video Retrieval from Reference Photos and Descriptions}

\author{
Anas Filali Razzouki$^{1,2}$ \quad Killian Steunou$^{1,2}$ \quad Khalil Guetari$^{2}$ \quad Thomas Kling$^{2}$\\
Yannis Tevissen$^{1,2}$ \quad Moun\^{\i}m El-Yacoubi$^{1}$\\[0.6em]
{\normalsize $^{1}$T\'el\'ecom SudParis, Institut Polytechnique de Paris, France}\\
{\normalsize $^{2}$Moments Lab Research, France}
}

\begin{document}



\maketitle
    
\providecommand{\upd}[1]{#1}

\begin{abstract}
Retrieving video shots in which specific people, identified by reference photos, perform particular actions requires jointly reasoning about identity and action. While existing person-centric retrieval benchmarks focus on single-person queries, retrieving interactions between multiple referenced people remains largely unexplored. We introduce CAST (Character-Aware Shot reTrieval), a benchmark for identity-conditioned video retrieval in which queries combine reference photographs with natural-language descriptions of actions or interactions. Built from LSMDC v2, CAST comprises 10{,}727 training queries and 783 test queries, including 195 multi-person queries, over a gallery of 5{,}210 test shots. Its hard negatives share either the queried identities or the described action, challenging models to account for both. We further investigate how multimodal embedding models can be adapted to this task by distilling a task-specific reranker and incorporating facial identity supervision during training. The resulting models retrieve shots through precomputed gallery embeddings without explicit face matching at inference. CAST-8B achieves 49.7\% mAP, outperforming the strongest face-based baseline by 13.3 percentage points, while CAST-2B achieves 43.3\% mAP at 66\,ms per query, 1.7$\times$ faster than this baseline. Our training more than doubles the mAP of the zero-shot embedding models. On the ITCPR composed person retrieval benchmark, our framework achieves state-of-the-art R@1 and mAP with reranking. We will publicly release the benchmark, associated data, and trained models.
\end{abstract}

\section{Introduction}
\label{sec:intro}

Editorial teams in film and television routinely search their archives for the shots in which a specific person performs a specific action.
In practice, this search is mostly textual: a query such as a name and an action is matched against manual logs or metadata.
\emph{Identity-conditioned shot retrieval}, in which a query combines a reference photograph of each person with a natural-language description of what they do and the matching shots are retrieved from a gallery, remains little explored.
Yet it would let editorial teams search footage that has never been annotated, such as new rushes or archives without shot-level logs, and find people by their appearance even when their names are missing from the metadata.
The system must recognize the referenced person and the described action in the same shot; a shot showing only one of them does not match.
The difficulty grows when several people are referenced through separate photographs and the description involves their interaction.
Given photographs of two people and the description ``Person~1 hugs Person~2'', a shot in which both people appear but do something else does not match the query, nor does a shot in which Person~1 hugs someone else while Person~2 is present.
We therefore define the multi-person task as \emph{interaction-presence retrieval}: a shot matches if all referenced people appear and take part in the described interaction.

Person-centric video retrieval has been studied under several formulations.
The TRECVID instance search task~\cite{awad2020trecvid} introduced person--action search in 2019: systems retrieve shots of a person, identified by reference images, performing one of a set of predefined actions.
CLIP-PAD~\cite{korbar2022personalised} introduced the CiA benchmark, which replaces predefined actions with free-form descriptions paired with a reference face.
More recently, FAFA~\cite{fafa} introduced ITCPR, where a reference photograph and a text retrieve images of the same person under a different appearance.
All of these benchmarks reference a single person per query; none evaluates queries that reference several people through separate photographs and describe their interaction.

Early person--action retrieval systems combine explicit face recognition with action recognition~\cite{longxiang2020whu,zhao2019instance,yang2021instance}.
They score identity by matching the reference faces against the faces detected in each shot, score the action with action recognition models, and fuse or rerank the two scores, sometimes with person tracking, pose estimation, or human--object interaction detection to attach actions to people.
Running several recognition models on every shot makes indexing expensive, and detection or tracking errors propagate to the ranking.
CLIP-PAD~\cite{korbar2022personalised} removes face matching from the gallery side by mapping the reference face embedding to a learnable text token of a CLIP model, so that shots are retrieved by similarity with precomputed video embeddings; it is, however, designed for a single person per query.

Vision--language models now enable efficient retrieval through shared embeddings of queries and videos.
VideoCLIP~\cite{xu2021videoclip} aligns video and text through contrastive learning with hard negatives, and Video-ColBERT~\cite{reddy2025video} adds fine-grained spatial and temporal interactions.
Qwen3-VL-Embedding and Qwen3-VL-Reranker~\cite{li2026qwen3} combine multimodal embedding learning with a cross-encoder reranker, which reads the query and the candidate jointly, and use knowledge distillation between the two.
These general-purpose models are not designed for identity-conditioned retrieval, which remains underexplored, yet they are well suited to it: they encode photographs and text jointly in one query, handle multi-frame video, and pair an embedding model with a reranker that can serve as its teacher.

We address these limitations with a benchmark and a method.
The \emph{CAST} (\textbf{C}haracter-\textbf{A}ware \textbf{S}hot re\textbf{T}rieval) \emph{benchmark} contains identity-conditioned queries over movie shots, each with one or more reference photographs and a description of an action or interaction.
Starting from identity-annotated movie shots, a vision--language model writes queries grounded in the identified actors, finds further matching shots, and selects hard negatives in which either the identities or the action differ from the query.
The test set contains 783 queries over 5{,}210 shots from 12 movies, 195 of them multi-person.
It provides an alternative to the single-person benchmarks of TRECVID and CiA, whose data are no longer readily accessible, and extends the task to interaction-presence queries over several referenced people.

Alongside the benchmark, we provide the \emph{CAST framework}, a retrieval framework that keeps the efficiency of an embedding model while learning the identity-conditioned judgments of a reranker.
We fine-tune Qwen3-VL-Reranker-8B to judge whether a shot matches the reference photographs and the description, distill its judgments into Qwen3-VL-Embedding, and add a training-only loss that aligns the shot embeddings with a face recognizer (ArcFace~\cite{deng2019arcface}).
Our controlled ablations show what each component brings to identity-conditioned retrieval: adapting the vision encoder and distilling a task-specific teacher account for most of the gain, teacher-selected negatives matter mainly for multi-person queries, and the identity loss adds a small but consistent improvement.
At search time, the gallery is encoded once, and a query is a single forward pass followed by a dot product, without face detection, tracking, or identity matching.

In summary, our contributions are:
\begin{itemize}
  \item the CAST benchmark for identity-conditioned shot retrieval, with single-person queries and multi-person interaction-presence queries, hard negatives that share either the identities or the action of each query, and a blind human audit of its test labels;
  \item CAST, an empirically validated framework for identity-conditioned retrieval that distills a fine-tuned reranker into an embedding model aligned with a face recognizer during training only, so that search requires no face pipeline, with controlled ablations of each component;
  \item an evaluation of accuracy and cost against face-based and embedding baselines, evidence that the accuracy of CAST tracks facial recognizability, and a transfer study on ITCPR, where CAST sets a new state of the art in R@1 and mAP with a reranker.
\end{itemize}

\section{CAST Benchmark}
\label{sec:benchmark}

\subsection{Source Data}

We build our dataset from the LSMDC v2 collection~\cite{lsmdc,lsmdc2016MovieAnnotationRetrieval,maharaj2017dataset}, downloading 83 movies.
Following the BAC pipeline~\cite{razzouki2026beyond}, we collect five reference face images per actor from IMDb cast lists and web images, filtered with InsightFace, then match the faces detected in each shot to these references and track them across frames.
Each shot is represented by eight sampled frames with the boxes of its identified actors.
Splits are made at the movie level: 5{,}210 test shots from 12 movies, and 31{,}890 training shots from 71 other movies, 7 of which are held out for validation.

\subsection{Query Construction}

\begin{figure*}[t]
  \centering
  \includegraphics[width=0.89\textwidth]{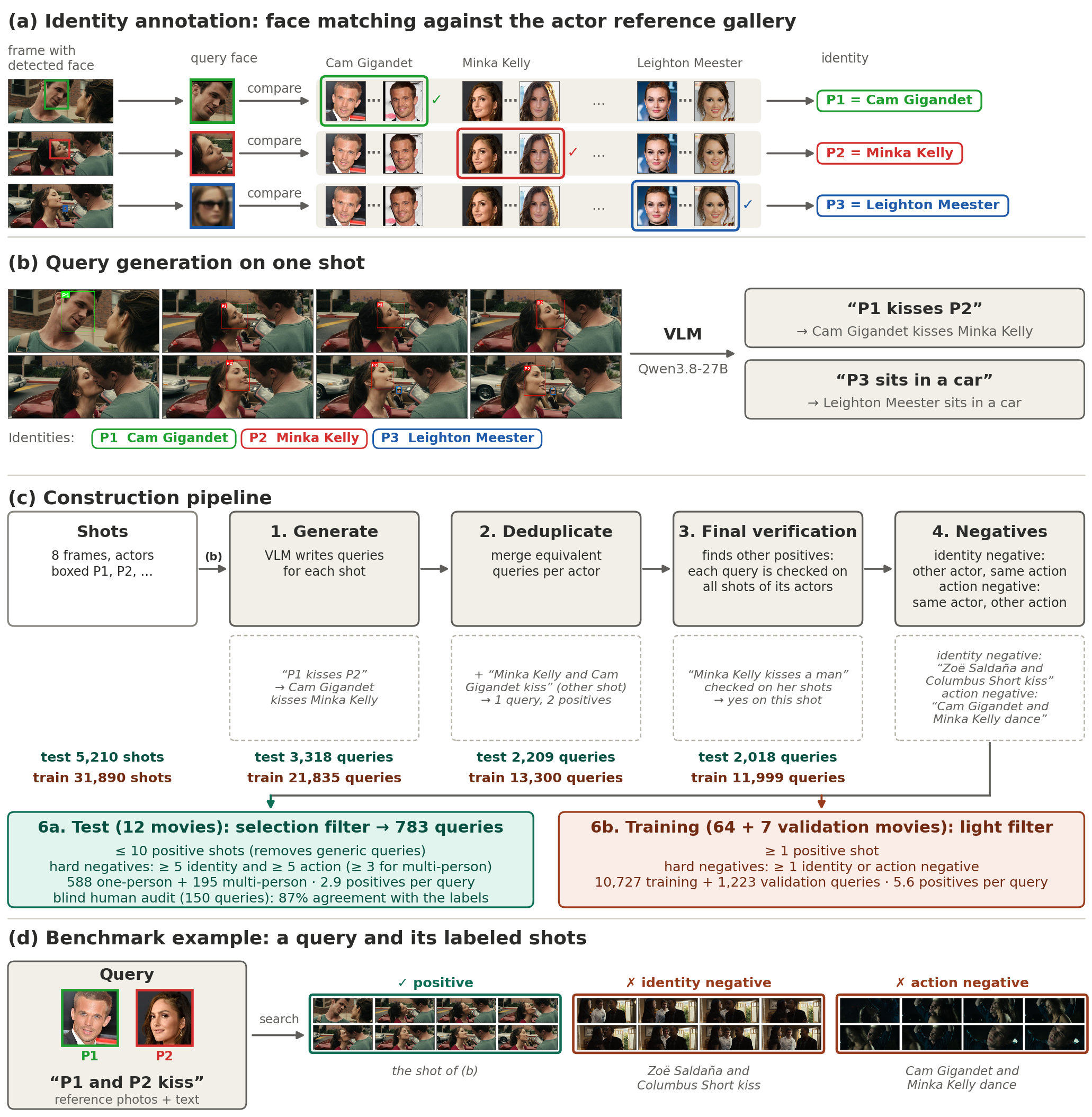}
  \caption{\textbf{Query construction.}
  (a)~For each shot, the VLM sees the eight frames with every identified actor boxed (P1, P2, \dots) and writes short queries about these people.
  (b)~Queries are merged across shots, verified on all shots of their actors to find further positives, and paired with identity and action hard negatives. The test split keeps only selective queries with enough hard negatives; the training split uses a lighter filter.}
  \label{fig:data_pipeline}
\end{figure*}

As shown in \cref{fig:data_pipeline}, a vision--language model (Qwen3.8-27B) receives the eight frames of a shot, with each identified actor boxed and labeled (P1, P2, \dots), and writes queries describing an individual action, a spatial relation, or an interaction between the labeled people~\cite{razzouki2026beyond}.
The source shot is the first positive of each query.
Equivalent queries on the same identities are merged and their positives pooled.
The model then checks each query against the other shots that contain its actors, adding further positives and discarding uncertain matches.
Finally, each query receives one reference photograph per referenced actor, drawn from that actor's five reference images; the query text never names the actors.

\subsection{Hard Negatives and Splits}

Each query has two types of hard negatives: \emph{identity negatives}, in which other people perform the queried action, and \emph{action negatives}, in which the referenced people appear but do something else.
The test split keeps queries with at most 10 positives, at least 5 identity negatives, and at least 5 action negatives (3 for multi-person queries).
A query can therefore be solved neither from its description alone nor from its reference photographs alone.
The test set contains 783 queries (588 single-person, 195 multi-person) over 5{,}210 shots, with 2.9 positives per query on average.
A lighter filter yields 10{,}727 training queries from 64 movies (CAST-train) and 1{,}223 validation queries from 7 movies.

\subsection{Human Validation}
\label{sec:audit}

Five annotators audited \upd{150} test queries (75 single- and 75 multi-person) and 1{,}337 of their shots: all positives, one identity and one action negative, and the two highest-ranked negatives of CAST-8B and of Late Fusion. The audit was blind: annotators judged whether each shot matches without seeing its benchmark label (protocol in \cref{sec:supp_audit}).
To measure how consistent the annotators are with each other, five queries (45 shots) were labeled by all five; on the 42 shots that no annotator marked unsure, any two annotators gave the same answer on 96.7\% of shots (Fleiss' $\kappa=0.93$).
The annotators' answers agree with the benchmark labels on 86.6\% of the audited shots, which over-represent hard cases (\cref{sec:supp_audit}).
Re-scoring the audited queries with the human labels leaves the ranking of the methods unchanged: CAST-8B 60.0, CAST-2B 48.5, and Late Fusion 41.7 mAP.

\section{Method}
\label{sec:method}

\begin{figure*}[t]
  \centering
  \includegraphics[width=1.0\textwidth]{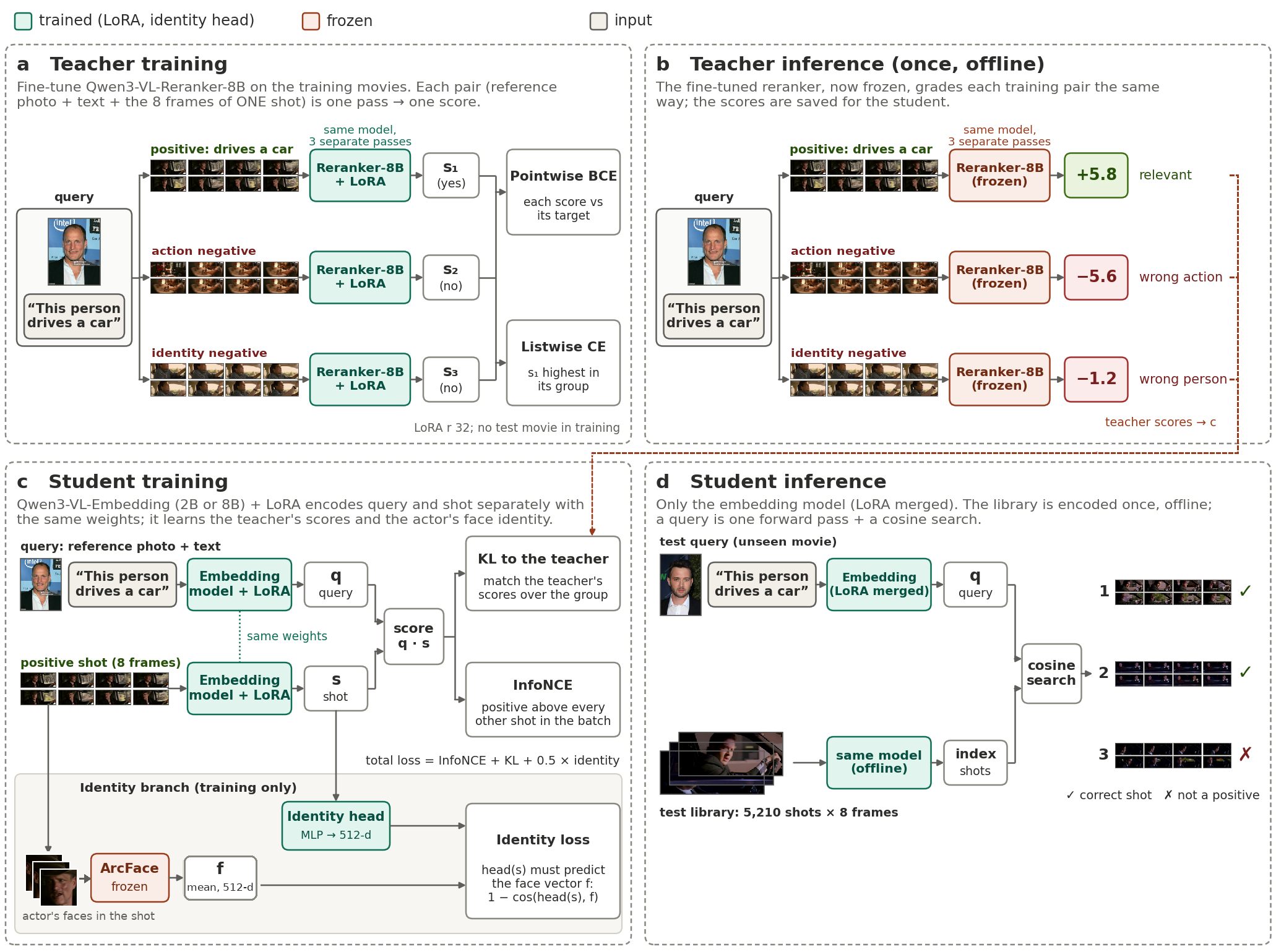}
  \caption{\textbf{Teacher--student training.}
  (a)~A Qwen3-VL-Reranker-8B teacher is fine-tuned on (query, shot) pairs with pointwise and listwise losses.
  (b)~The frozen teacher scores the training shots once, offline.
  (c)~The Qwen3-VL-Embedding student encodes query and shot separately and is trained with InfoNCE, distillation of the teacher scores, and an identity loss that is removed at search time.
  (d)~At test time, the gallery is encoded once and a query is a single forward pass plus a cosine search.}
  \label{fig:method}
\end{figure*}

CAST combines a reranker teacher, teacher-guided embedding learning, and facial identity alignment (\cref{fig:method}).
We describe how each component is adapted to identity-conditioned retrieval.
The reranker learns to separate shots that match the referenced people and their action from shots with an identity or action mismatch; its judgments are distilled into an embedding model, which is additionally aligned with a face recognizer during training.

\subsection{Problem Formulation}

A query $q=(\{r_i\}_{i=1}^{m},t)$ consists of one reference photograph $r_i$ for each of the $m$ referenced people and a description $t$; a shot $v$ consists of eight uniformly sampled frames.
An encoder $f$ (Qwen3-VL-Embedding) maps queries and shots independently into a shared space, and shots are ranked by
\begin{equation}
  s(q,v)=\frac{f(q)^{\top}f(v)}{\lVert f(q)\rVert\,\lVert f(v)\rVert}.
  \label{eq:score}
\end{equation}
Embeddings are $\ell_2$-normalized, so \cref{eq:score} is a dot product. Since $f(v)$ does not depend on the query, the gallery is encoded once and reused for every query.

\subsection{Reranker Teacher}

A cross-encoder reads the query and the shot together and can compare each referenced face with the people in the shot, but it requires one forward pass per query--shot pair, which is too slow to search a large gallery.
We therefore fine-tune Qwen3-VL-Reranker-8B on CAST-train and use it only as a teacher.
Each training group contains one positive, three identity negatives, three action negatives, and one random negative, and the reranker is trained with a pointwise binary cross-entropy and a listwise cross-entropy over the group.

\subsection{Teacher-Guided Embedding}

For each training query, we build a pool of up to 32 candidate shots: up to four positives, four identity negatives, four action negatives, and 20 shots retrieved by the zero-shot Qwen3-VL-Embedding-8B.
The frozen teacher scores every pool once, offline.
Its scores discard likely false negatives and select, for each query, the two negatives it ranks highest; these form a training group with the positive and one random negative.
Qwen3-VL-Embedding is itself distilled from a general-purpose reranker~\cite{li2026qwen3}; here, the teacher is first fine-tuned to judge whether a shot shows the referenced people performing the described action, and its scores serve both as soft targets and to select and filter the negatives of each query.
The student is trained with
\begin{equation}
  \mathcal{L}=\mathcal{L}_{\text{NCE}}+\mathcal{L}_{\text{KD}}+\lambda\,\mathcal{L}_{\text{id}},
  \label{eq:loss}
\end{equation}
where $\mathcal{L}_{\text{NCE}}$ is the InfoNCE loss~\cite{oord2018representation} over all shots in the batch, $\mathcal{L}_{\text{KD}}$ is the KL divergence between the teacher's and the student's score distributions over the query's group, and $\mathcal{L}_{\text{id}}$ is the identity loss below.
Both the language model and the vision encoder are adapted with LoRA~\cite{hu2022lora}.

\subsection{Identity Alignment and Inference}

The contrastive and distillation losses supervise the ranking, but not explicitly who appears in a shot.
We add a training-only identity loss, applied to single-person training queries.
A frozen ArcFace model embeds the queried actor's face in the positive shot $v^{+}$, averaged over the frames and tracks in which this actor is detected and $\ell_2$-normalized, giving a target $a$.
A lightweight projection head $h$ maps the student's shot embedding into the ArcFace space:
\begin{equation}
  \mathcal{L}_{\text{id}}=1-\cos\!\big(h(f(v^{+})),\,a\big).
  \label{eq:id}
\end{equation}
At inference, ArcFace and $h$ are discarded. The gallery is encoded once, and a query requires one forward pass and a dot product (\cref{eq:score}), with no face detection, tracking, or pairwise query--shot processing.

\section{Experiments}
\label{sec:experiments}

\subsection{Experimental Setup}
\label{sec:setup}

\textbf{Protocol.}
We evaluate on the CAST test set: 783 queries ranked against all 5{,}210 shots of 12 movies unseen during training, with one reference photograph per person.
Our main metric is mAP, which accounts for every positive of a query.
We also report R@1, the fraction of queries whose top-ranked shot is correct, and R@5 and R@10, the fraction of a query's positives retrieved in the top 5 and top 10.
Results are given overall and for single- and multi-person queries, and differences are tested with a two-sided paired bootstrap over queries, for which we report $p$-values.
All timings use a single NVIDIA A10G GPU.

We compare CAST with two groups of baselines: zero-shot models, and models trained on CAST-train, which separate the benefit of our data from that of our training.

\textbf{Zero-shot models.}
SigLIP2~\cite{tschannen2025siglip2} scores the reference photographs and the description against the video frames separately and combines the normalized scores.
Qwen3-VL-Embedding-2B/8B encode the photographs and the text jointly; they are the models from which CAST is initialized.

\textbf{Trained models.}
CLIP-PAD~\cite{korbar2022personalised} is retrained on CAST-train following its original procedure.
SigLIP2 is fine-tuned with LoRA adapters on its image and text encoders and a lightweight head that fuses the photograph and description embeddings into one query vector.
\emph{Scene--Identity Late Fusion} is a modern version of the face-based person--action pipelines~\cite{longxiang2020whu,zhao2019instance,yang2021instance}, which score identity by face matching and the action separately, then fuse the two scores: it adds to the text--video similarity of the fine-tuned SigLIP2 an identity score obtained by matching the reference faces with ArcFace to the faces detected and tracked in each shot. Both scores are standardized over the gallery and summed with equal weights (\cref{sec:supp_latefusion} illustrates the pipeline).
Finally, CAST without the teacher keeps the same architecture, data, and identity loss.

\subsection{Implementation Details}
\label{sec:impl}

CAST fine-tunes Qwen3-VL-Embedding-2B and -8B with LoRA (rank 32, $\alpha=32$, dropout 0.05) on the language model and the vision encoder, with learning rate $2\times10^{-5}$, distillation weight 1, and identity weight $\lambda=0.5$.
Validation mAP is used for early stopping and to select the LoRA configuration, the learning rate, $\lambda$, and the input resolution.
Each shot is encoded from eight frames at native resolution (about 2{,}934 visual tokens), and each reference photograph is capped at 261 visual tokens: on validation, reducing the photograph resolution leaves mAP unchanged, whereas reducing the shot resolution lowers it for both model sizes.

\subsection{Results}
\label{sec:results}

\subsubsection{Accuracy and Search Latency}
\label{sec:res_latency}

\Cref{fig:latency}a compares accuracy and query latency.
CAST is the most accurate method and among the fastest: CAST-2B reaches 43.3 mAP at 66\,ms per query, and CAST-8B 49.7 mAP at 141\,ms.
The strongest baseline, Scene--Identity Late Fusion, reaches 36.4 mAP at 111\,ms.
This baseline method must match each reference face against the face tracks of every shot, which grows with the number of people: multi-person queries take 365\,ms, against 102\,ms for CAST-2B and 218\,ms for CAST-8B.
CAST-2B is thus 1.7$\times$ faster than Late Fusion and 6.9 mAP more accurate, and CAST-8B is 13.3 mAP more accurate for 30\,ms more per query.
Both gains are significant ($p<0.001$), and CAST-2B and CAST-8B are better than Late Fusion on 11 of the 12 test movies.
SigLIP2 trained on CAST-train is faster (58\,ms) but far less accurate (26.0 mAP).
The zero-shot Qwen3-VL-Embedding models have the same cost as CAST but less than half its accuracy (17.2 and 23.0 mAP): the gains come from our training, not from the backbone.
Finally, splits share no movie, but 22 of the 94 test actors also appear in training movies. CAST is not more accurate on their queries: 45.4 \vs 50.0 mAP for CAST-8B and 42.9 \vs 43.2 for CAST-2B on seen \vs unseen actors (135 \vs 610 queries; 38 queries mixing both are excluded), differences that are not significant ($p=0.16$ and $0.92$, permutation test).
The lower score on seen actors suggests harder queries rather than an effect of training: Late Fusion, which does not learn these actors, shows a similar drop (33.8 \vs 36.1).

\subsubsection{Indexing Cost}
\label{sec:res_index}

\begin{figure*}[t]
  \centering
  \includegraphics[width=\linewidth]{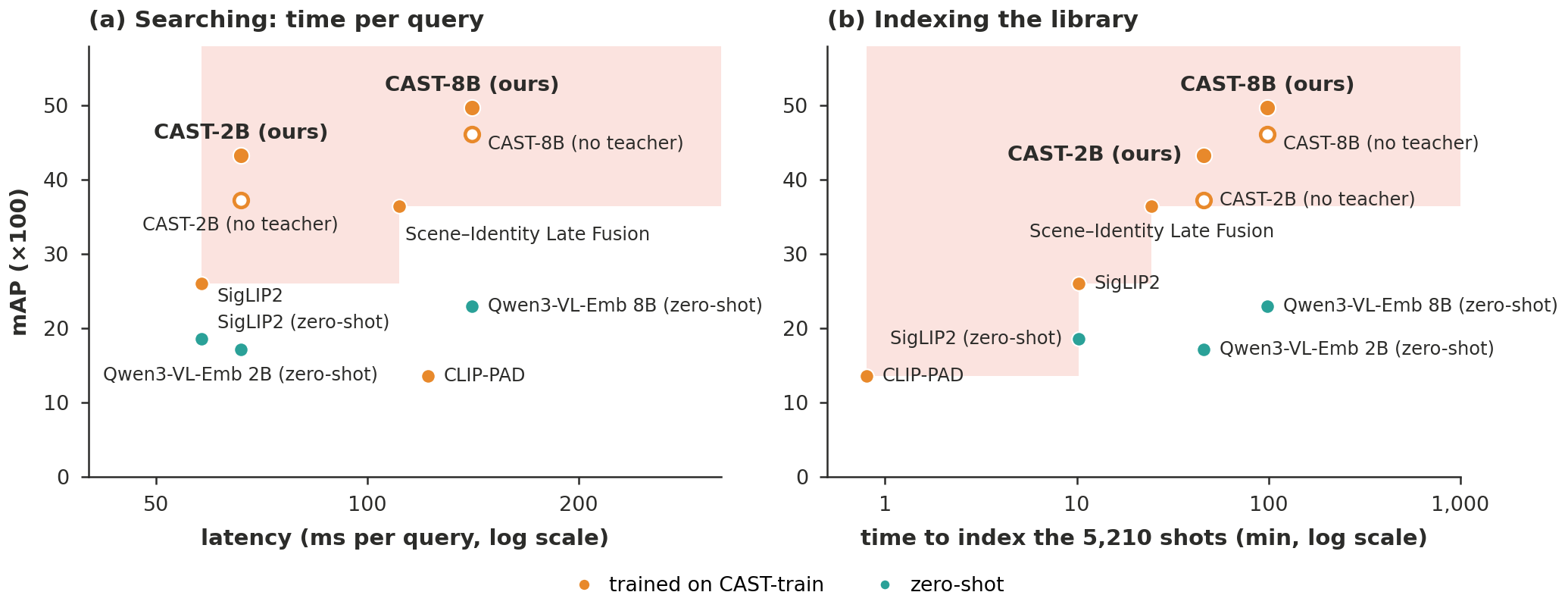}
  \caption{\textbf{Accuracy \vs cost} on the CAST test set (783 queries, 5{,}210 shots, one A10G GPU).
  (a)~Query latency, from the raw reference photograph to the top-100 shots, averaged over single- and multi-person queries in benchmark proportions.
  (b)~Time to index the gallery.
  Orange: trained on CAST-train; teal: zero-shot; hollow: CAST without teacher.}
  \label{fig:latency}
\end{figure*}

Indexing is where CAST pays for its accuracy (\cref{fig:latency}b): encoding the 5{,}210 shots takes 46\,min for CAST-2B and 98\,min for CAST-8B, against 24.5\,min for Late Fusion, even though Late Fusion runs face detection, ArcFace, and tracking on every frame.

\begin{figure}[t]
  \centering
  \includegraphics[width=\linewidth]{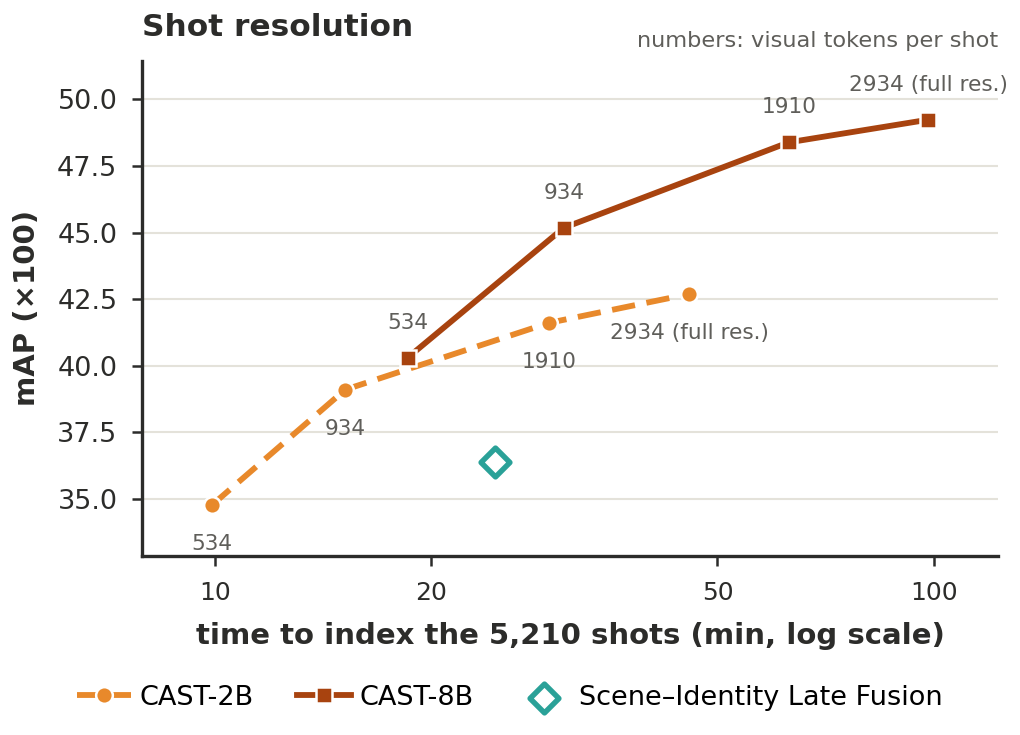}
  \caption{\textbf{Shot resolution \vs indexing time} (test set, mAP), with reference photographs fixed at 261 tokens.
  Numbers give the visual tokens per shot; ``full res.'' is the native resolution used by CAST.
  The diamond marks Scene--Identity Late Fusion at its own indexing cost.}
  \label{fig:resolution}
\end{figure}

To understand where this cost comes from, we vary the number of visual tokens per shot (\cref{fig:resolution}).
Indexing time falls almost linearly with the number of tokens.
With 1{,}910 tokens per shot, indexing is 36\% faster (29\,min for CAST-2B, 63\,min for CAST-8B) at a cost of only 1.1 and 0.8 mAP.
At a budget comparable to Late Fusion, CAST is both faster and more accurate: CAST-2B with 934 tokens indexes the gallery in 15\,min at 39.1 mAP, and CAST-8B with 534 tokens in 19\,min at 40.3 mAP, against 24.5\,min and 36.4 mAP for Late Fusion.
Lower resolutions hurt multi-person queries most: at 934 tokens, CAST-8B drops from 51.3 to 44.3 mAP on them, since telling several people apart in a shot requires fine visual detail.
The indexing cost of CAST is therefore a choice made for accuracy, and it can be traded for speed when needed.

\subsubsection{Contribution of Each Component}
\label{sec:res_ablation}

\Cref{tab:ablation} removes one component of CAST-2B at a time.
Training is essential: the zero-shot model reaches 17.2 mAP, and CAST 43.3.
Adapting the vision encoder in addition to the language model matters most: without it, mAP drops by 6.1 points ($p<0.001$), on single-person (42.9 $\to$ 37.5) and multi-person queries (44.6 $\to$ 36.5) alike. Recognizing a given actor and a fine-grained action in movie frames requires adapting the visual features, not only the language model.
Removing the teacher has a similar overall effect (6.0 mAP lower, $p<0.001$), but concentrated on multi-person queries, which fall from 44.6 to 32.7 while single-person queries lose 4.0 points.
This variant, trained with InfoNCE and the identity loss on the labeled negatives, is an ordinary supervised contrastive baseline.
The teacher serves two purposes, soft targets and the selection of negatives, which we separate with a variant that keeps the distillation loss but replaces the teacher-selected and filtered negatives with random labeled ones.
Distillation alone recovers 4.9 of the 6.0 points (37.3 $\to$ 42.2, $p<0.001$), on single- (+4.1) and multi-person queries (+7.2): the teacher grades how well each shot matches, which binary labels do not.
Teacher-selected negatives add 1.1 mAP overall, which is not significant ($p=0.09$), but 4.7 on multi-person queries (39.9 $\to$ 44.6, $p=0.001$): random negatives rarely contain the same people performing another action, and the filter removes the 19\% of candidate negatives that the teacher judges to match the query.
Finally, the identity loss adds 1.2 mAP ($p<0.001$): a small but consistent gain at no inference cost.

\begin{table*}[t]
  \centering
  \resizebox{\textwidth}{!}{%
  \begin{tabular}{l cccc cccc cc c}
    \toprule
    & \multicolumn{2}{c}{Teacher} & & & \multicolumn{4}{c}{All queries} & \multicolumn{2}{c}{mAP by query type} & \\
    \cmidrule(lr){2-3} \cmidrule(lr){6-9} \cmidrule(lr){10-11}
    Variant & KD & Negatives & ID loss & Vision LoRA & mAP & R@1 & R@5 & R@10 & Single & Multi & $\Delta$ mAP \\
    \midrule
    \textbf{CAST-2B (final)} & \checkmark & \checkmark & \checkmark & \checkmark & \textbf{43.3} & \textbf{44.7} & \textbf{48.6} & \textbf{60.9} & 42.9 & \textbf{44.6} & -- \\
    No identity loss & \checkmark & \checkmark & $\times$ & \checkmark & 42.1 & 42.4 & 47.8 & 60.1 & 41.6 & 43.6 & $-$1.2$^{\dagger}$ \\
    Random negatives & \checkmark & $\times$ & \checkmark & \checkmark & 42.2 & 42.9 & 48.1 & 58.6 & \textbf{43.0} & 39.9 & $-$1.1 \\
    No teacher & $\times$ & $\times$ & \checkmark & \checkmark & 37.3 & 38.6 & 41.5 & 52.7 & 38.9 & 32.7 & $-$6.0$^{\dagger}$ \\
    No vision LoRA & \checkmark & \checkmark & \checkmark & $\times$ & 37.2 & 34.9 & 41.8 & 55.8 & 37.5 & 36.5 & $-$6.1$^{\dagger}$ \\
    Zero-shot & $\times$ & $\times$ & $\times$ & $\times$ & 17.2 & 13.7 & 20.2 & 28.1 & 19.6 & 9.8 & $-$26.1 \\
    \bottomrule
  \end{tabular}}
  \caption{\textbf{Ablation of CAST-2B} on the test set. All metrics in \%. Single and Multi give the mAP on single- and multi-person queries; $\Delta$ mAP is the difference with the final model.
  KD: distillation of the teacher scores; Negatives: negatives selected and filtered by the teacher (otherwise random labeled negatives).
  $^{\dagger}$$p<0.001$, paired bootstrap over queries.}
  \label{tab:ablation}
\end{table*}

\subsubsection{Does CAST Learn to Recognize Faces?}
\label{sec:res_identity}

\begin{figure}[t]
  \centering
  \includegraphics[width=0.85\linewidth]{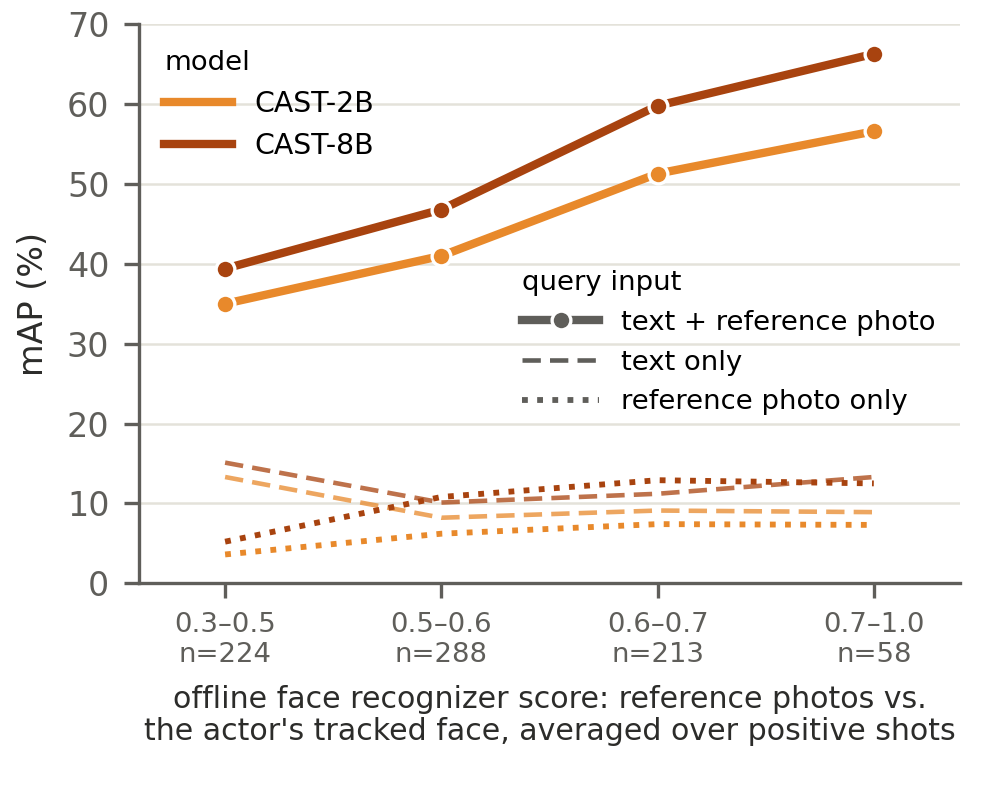}
  \caption{\textbf{Accuracy \vs how recognizable the actor is}: queries grouped by ArcFace score.}
  \label{fig:identity}
\end{figure}

For each test query, we measure offline how recognizable the queried actor is: a face recognizer (ArcFace) compares the actor's reference photographs with the actor's tracked faces in each positive shot, and we average this score over the positive shots, taking the lower of the two actors for multi-person queries.
\Cref{fig:identity} groups the queries by this score and evaluates the same model with three inputs: text and reference photograph, text only, and reference photograph only.

With both inputs, the accuracy of CAST follows the face recognizer's score: the better the reference photographs match the faces in the positive shots, the higher the mAP (35.0 $\to$ 56.6 for CAST-2B, 39.4 $\to$ 66.3 for CAST-8B).
Without detecting, cropping, or matching any face, the CAST embedding is thus sensitive to facial identity in the same way as a face recognizer.
This behavior comes from the reference photograph: with text only, mAP stays between 8 and 15 and does not follow the score, so clearer faces do not make shots easier by themselves. The photograph alone stays between 4 and 13 mAP.
Only the combination works: for CAST-8B in the 0.6--0.7 range, the two inputs reach 59.8 mAP together, against 11.2 and 12.9 separately.

\begin{figure}[t]
  \centering
  \includegraphics[width=0.85\linewidth]{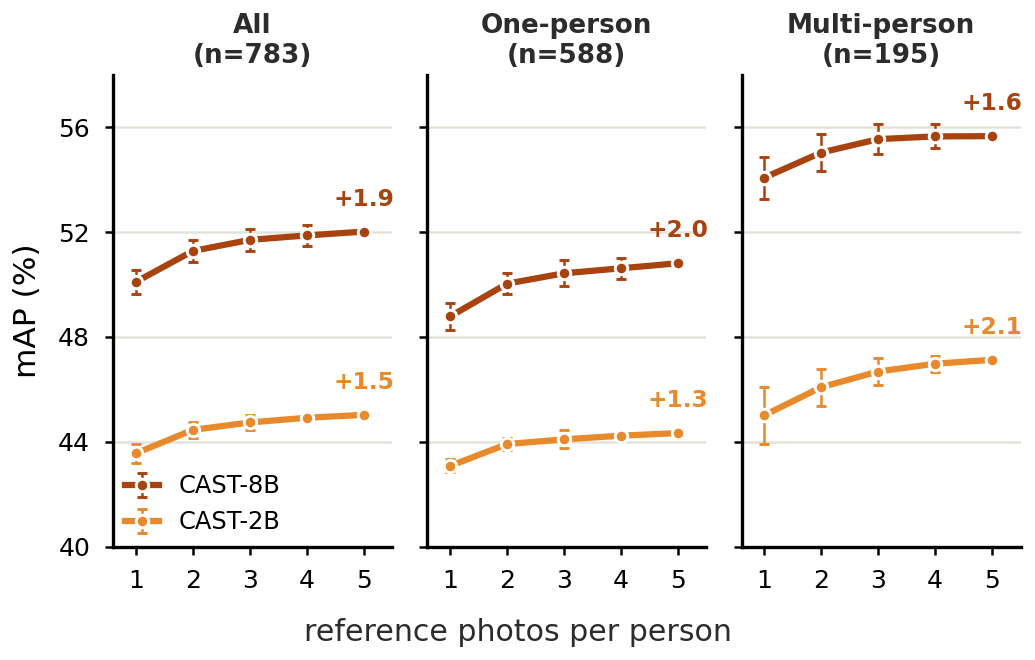}
  \caption{\textbf{Number of reference photographs per person} at test time: mAP averaged over all choices of $k$ among five (error bars: std.); numbers give the gain from one to five.}
  \label{fig:refcount}
\end{figure}

So far, each person is given by a single reference photograph. Since up to five photographs are available per person, we test the effect of adding more: averaging the query embeddings over these photographs at test time raises mAP by 1.5 for CAST-2B and 1.9 for CAST-8B, without retraining (\cref{fig:refcount}).

\subsubsection{Roles within an Interaction}
\label{sec:res_binding}

Beyond the presence of the referenced people, a model should also know who does what, \eg, ``Person~1 pushes Person~2'' rather than ``Person~2 pushes Person~1''.
The benchmark has no shots labeled with the reversed roles, so we reverse the query instead: on the 68 test queries whose meaning depends on the order, we swap the two reference photographs and check whether the positive shots score higher with the photographs in the correct order.
CAST does so on only 37 of the 68 queries, which is chance level ($p=0.54$), whereas the fine-tuned teacher does so on 62 when instructed that the order of the photographs defines the roles (56 otherwise; \cref{sec:supp_binding}).
The same instruction does not help CAST (37 and 34 of 68 for CAST-2B and CAST-8B).
Our claims on multi-person queries are therefore limited to interaction-presence retrieval.

\subsection{Transfer to Composed Person Retrieval}
\label{sec:itcpr}

To test whether the CAST framework transfers to another task, we apply it to composed person retrieval (CPR)~\cite{fafa}, a related public task on still images: a reference image of a person and a text describing how this person appears in the target image (\eg, a change of clothes) must retrieve the gallery images of the same person that match the description.
As in our benchmark, a query is solved only by combining \emph{who} and \emph{what}.
We train CAST with the same pipeline as on movies (\cref{sec:method}) on the synthetic SynCPR triplets~\cite{fafa} and evaluate on ITCPR~\cite{fafa}, a manually annotated test set with 2{,}202 queries and 20{,}510 gallery images never seen in training.
As a single embedding model, CAST-8B ranks second in R@1 (43.5), behind FAFA (46.54), and ahead of SPRC and CaLa trained on the same data, although SPRC remains better in R@5, R@10, and mAP (\cref{tab:itcpr}).
Reranking the top 10 images of CAST sets a new state of the art in R@1 and mAP: 51.2 R@1 and 56.3 mAP for CAST-8B, +4.7 R@1 and +0.7 mAP over FAFA, which remains better at R@5 and R@10, bounded by the top 10 of the first stage.
On one A10G GPU, CAST-2B indexes the 20{,}510 gallery images 1.7$\times$ faster than FAFA (174 \vs 291\,s) and answers a query as fast (49\,ms for both); CAST-8B takes 409\,s to index and 93\,ms per query.
Reranking the top 10 adds 0.32\,s per query, so the two-stage pipeline answers in about 0.4\,s, 8$\times$ slower than FAFA but more accurate.
CAST thus offers either a single embedding as fast as the state of the art, or a two-stage pipeline that surpasses it at a moderate query cost.

\begin{table}[t]
  \centering
  \resizebox{\linewidth}{!}{%
  \begin{tabular}{llcccc}
    \toprule
    Method & Training data & R@1 & R@5 & R@10 & mAP \\
    \midrule
    \textbf{CAST-8B + reranker (top 10)} & SynCPR & \textbf{51.2} & 62.0 & 63.7 & \textbf{56.3} \\
    \textbf{CAST-2B + reranker (top 10)} & SynCPR & 50.6 & 59.9 & 61.1 & 55.2 \\
    FAFA~\cite{fafa} (NeurIPS'25) & SynCPR & 46.54 & \textbf{66.21} & \textbf{73.12} & 55.60 \\
    \textbf{CAST-8B} & SynCPR & 43.5 & 58.6 & 63.7 & 50.6 \\
    SPRC~\cite{sprc} (ICLR'24) & SynCPR & 42.27 & 61.81 & 69.35 & 51.62 \\
    CaLa~\cite{cala} (SIGIR'24) & SynCPR & 39.33 & 60.85 & 68.66 & 49.29 \\
    \textbf{CAST-2B} & SynCPR & 38.6 & 54.8 & 61.1 & 46.3 \\
    FIRe2~\cite{fire2} + RaSa~\cite{rasa} & -- & 32.89 & 54.27 & 62.03 & 42.16 \\
    RDE~\cite{rde} (CVPR'24) & CUHK-PEDES & 29.79 & 51.82 & 60.49 & 40.10 \\
    RaSa~\cite{rasa} (IJCAI'23) & CUHK-PEDES & 28.02 & 49.23 & 57.77 & 38.04 \\
    CoVR-BLIP~\cite{covr} (AAAI'24) & WebVid-CoVR & 26.75 & 47.68 & 56.36 & 36.49 \\
    IRRA~\cite{irra} (CVPR'23) & CUHK-PEDES & 26.39 & 46.46 & 56.27 & 36.13 \\
    SPRC~\cite{sprc} (ICLR'24) & CIRR & 25.07 & 45.73 & 54.50 & 35.05 \\
    CaLa~\cite{cala} (SIGIR'24) & CIRR & 24.02 & 44.64 & 53.45 & 34.08 \\
    LinCIR~\cite{lincir} (CVPR'24) & -- & 23.93 & 44.46 & 53.18 & 33.95 \\
    Pic2Word~\cite{pic2word} (CVPR'23) & CC3M & 21.21 & 37.15 & 44.51 & 29.11 \\
    FIRe2~\cite{fire2} (TIFS'24) & LTCC & 10.76 & 22.84 & 29.29 & 17.00 \\
    CAL~\cite{cal} (CVPR'22) & LTCC & 9.86 & 22.34 & 29.20 & 16.45 \\
    SOLIDER~\cite{solider} (CVPR'23) & Market-1501 & 8.45 & 18.48 & 23.89 & 13.74 \\
    CLIP-ReID~\cite{clipreid} (AAAI'23) & Market-1501 & 7.95 & 18.12 & 22.75 & 13.31 \\
    TransReID~\cite{transreid} (ICCV'21) & Market-1501 & 7.27 & 17.30 & 22.75 & 12.57 \\
    \bottomrule
  \end{tabular}}
  \caption{\textbf{Full comparison on ITCPR} (2{,}202 queries, 20{,}510 gallery images), ranked by R@1. Prior results are taken from FAFA~\cite{fafa}. CAST is trained on SynCPR only; ``+ reranker'' reranks its top 10 with Qwen3-VL-Reranker-8B.}
  \label{tab:itcpr}
\end{table}

\section{Conclusion}
\label{sec:conclusion}

We introduced the CAST benchmark for identity-conditioned shot retrieval, whose hard negatives share either the identities or the action of each query, including multi-person interaction-presence queries.
Alongside it, the CAST framework shows empirically that distilling a task-specific reranker into an embedding model, with face-recognizer alignment during training only, is more accurate than face-based pipelines and general-purpose embeddings at a comparable or lower query cost, and transfers to composed person retrieval.

\textbf{Limitations.}
Labels are produced by a vision--language model, and only shots in which face matching found the referenced actors are judged; missed appearances count as negatives and lower absolute scores for all methods (\cref{sec:supp_audit}).
Our claims on multi-person queries are limited to interaction presence: CAST does not reliably distinguish the roles of the people within an interaction (\cref{sec:res_binding}).
Finally, the benchmark uses public images of professional actors; identity-conditioned retrieval of private individuals raises privacy concerns, and such models should not be deployed for surveillance.

\section*{LLM Usage Statement}
As part of the method, vision--language models generate and verify the benchmark queries (\cref{sec:benchmark}) and serve as the backbones of CAST and its teacher (\cref{sec:method}).
Separately, an LLM-based coding assistant (Claude, Anthropic) was used to help write analysis and plotting scripts and to edit the wording of the manuscript; the authors designed the study, checked all code and results, and take full responsibility for the content.

{
    \small
    \bibliographystyle{ieeenat_fullname}
    \bibliography{main}
}

\clearpage\typeout{SUPPLSTART=\the\value{page}}
\maketitlesupplementary
\appendix

\begin{figure*}[t]
  \centering
  \includegraphics[width=0.92\textwidth]{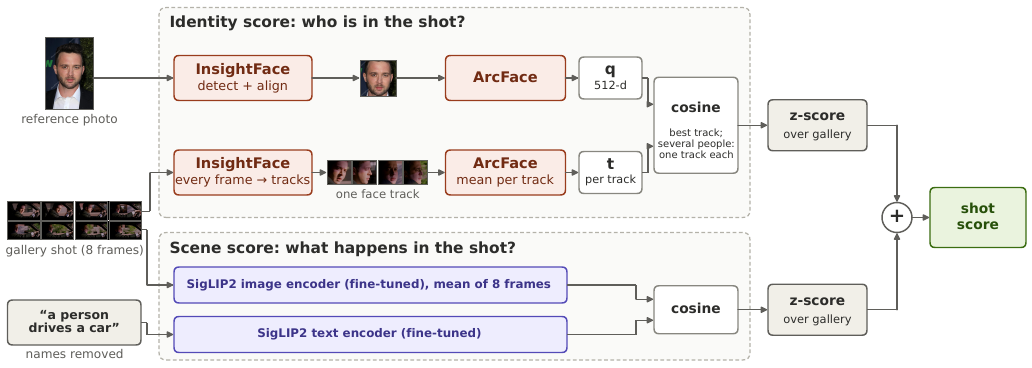}
  \caption{\textbf{Scene--Identity Late Fusion.} The identity score matches the reference face with the face tracks of the shot (InsightFace + ArcFace); the scene score compares the description with the frames (fine-tuned SigLIP2). Both are z-scored over the gallery and summed.}
  \label{fig:latefusion}
\end{figure*}

\section{Human Validation Protocol}
\label{sec:supp_audit}

Since queries, additional positives, and negatives are produced automatically, five annotators audited a sample of test queries, half single- and half multi-person, drawn across the 12 test movies.
For each query, they judged its positives, one identity negative, one action negative, and the two negatives ranked highest by CAST-8B and by Late Fusion.
The audit was blind: annotators saw the reference photographs, the description, and the eight frames of each shot, with the shots of a query in random order, but neither the benchmark label nor the type of the shot.
Each pair was labeled \emph{match}, \emph{no match}, or \emph{unsure}, with the reason for a mismatch (wrong person, wrong action, wrong partner, or swapped roles).
Four annotators labeled 30 queries each and one labeled 50, including five queries (45 shots) labeled by all five; each of the other 145 queries was labeled by a single annotator.
The human label of a shot is the answer of its annotator, or, on the five shared queries, the majority vote; shots with only \emph{unsure} answers or a tied vote are excluded from all statistics.

\paragraph{Agreement between annotators.}
On the 45 shared shots, 417 of the 438 pairs of answers that are not \emph{unsure} agree (95.2\%). Fleiss' $\kappa$ requires every annotator to answer each shot; on the 42 shots that no annotator marked \emph{unsure}, pairwise agreement is 96.7\% and $\kappa=0.93$.

\paragraph{Agreement with the benchmark labels.}
Over all audited shots, the annotators' answers agree with the benchmark labels on 86.6\% of the 1{,}483 judgments.
This figure is measured on the audited shots, which over-represent hard cases: positives, hard negatives, and the shots that two methods rank highest. Easy negatives, such as shots of other actors, which form most of the gallery, are not sampled, so this figure is not the accuracy of every gallery label.

\paragraph{Completeness of the labels.}
Relevance judgments are not exhaustive. Additional positives are searched only among the shots in which face matching found the referenced actors, as in pooled evaluation: a shot that matches the query but in which face matching missed an actor (\eg, a face turned away) remains a negative. This lowers absolute scores for all methods.

\paragraph{Re-scoring with the human labels.}
We re-score the 141 audited queries that have at least one human-confirmed positive (the other nine have none): positives rejected by the annotators are removed, audited shots they confirmed are added as positives, excluded shots are left out of the ranking, and unjudged gallery shots count as negatives.
\Cref{tab:audit_rescore} separates the effect of the query subset from that of the labels.
Both the subset and the corrected labels raise mAP, and the ranking of the methods is unchanged.
Since additional positives can only come from the top-ranked shots of CAST-8B and Late Fusion, these two methods gain more from the corrected labels than CAST-2B.

\begin{table}[h]
  \centering
  \small
  \begin{tabular}{l ccc}
    \toprule
    & All test queries & \multicolumn{2}{c}{141 audited queries} \\
    \cmidrule(lr){3-4}
    Method & benchmark labels & benchmark labels & human labels \\
    \midrule
    Late Fusion & 36.4 & 37.5 & 41.7 \\
    CAST-2B & 43.3 & 46.5 & 48.5 \\
    CAST-8B & 49.7 & 54.7 & 60.0 \\
    \bottomrule
  \end{tabular}
  \caption{\textbf{mAP with benchmark and human labels.} All test queries: 783.}
  \label{tab:audit_rescore}
\end{table}

\section{Scene--Identity Late Fusion Baseline}
\label{sec:supp_latefusion}

\Cref{fig:latefusion} shows how Late Fusion scores a gallery shot.
The identity score checks \emph{who} appears: InsightFace detects and aligns the faces in the reference photograph and in every frame of the shot, the detections of a shot are linked into face tracks, and ArcFace embeds every face; a track is represented by the mean ArcFace vector of its faces.
The identity score is the cosine similarity between the reference face and the best-matching track. For multi-person queries, each referenced person is assigned to a distinct track so as to maximize the total similarity, and a person without a track receives the lowest score of the query.
The scene score checks \emph{what} happens: SigLIP2, fine-tuned on CAST-train, compares the description, with the names removed, with the mean of the eight frame embeddings.
Each score is standardized over the gallery for the query, and the two are summed with equal weights.
Since face detection, ArcFace, and tracking run on every frame of the gallery, indexing takes 24.5\,min, and the assignment step makes multi-person queries slower (365\,ms).

\section{Roles within an Interaction: Details}
\label{sec:supp_binding}

We test the roles within an interaction by swapping the two reference photographs, which turns ``Person~1 hugs Person~2'' into ``Person~2 hugs Person~1'', and checking whether the positive shots score higher with the photographs in the correct order.
We restrict this test to the 68 test queries whose meaning changes with the order: 53 directed interactions and 15 asymmetric spatial relations (\eg, behind, between); most spatial relations, such as \emph{next to}, are symmetric.
The fine-tuned teacher, which reads the query and the shot jointly, prefers the correct order on 62 of the 68 queries when its instruction states that the order of the photographs defines the roles.
CAST-2B and CAST-8B do so on only 37 ($p=0.54$, binomial test): the role information is largely lost in distillation, since standard training never presents a reversed query.
A role-aware distillation variant raises this number from 37 to 41 of 68, a trend that is not significant on this test set (\cref{sec:supp_roles}).
Late Fusion is insensitive to the order by construction, since it assigns faces to tracks with the best total score.
CAST thus retrieves shots in which the referenced people take part in the described interaction, but it does not reliably distinguish their roles within it. Our claims on multi-person queries are therefore limited to interaction-presence retrieval, and role-sensitive retrieval is left to future work.

\section{Role-Aware Distillation}
\label{sec:supp_roles}

We therefore train a variant of CAST-2B that distills the teacher's order preference directly: for the 670 order-sensitive training queries, the student's scores of the positive shot under the correct and the swapped order are matched to the teacher's with a KL divergence.
This role-aware distillation raises the number of queries where the correct order wins from 37 to 41 of 68 (60\%), and from 34 to 39 when each shot is encoded from its central frame only, without a significant change in mAP (42.9 against 43.3, $p=0.20$).
The trend is consistent across both encodings, but the test set is small: the gain is not significant on 68 queries ($p=0.11$ against chance), and confirming an effect of this size would require about 190 order-sensitive queries (80\% power at the 5\% level).

\section{Number of Reference Photographs}
\label{sec:res_refcount}

All results so far use one reference photograph per person. Since the benchmark provides five per actor, we test whether CAST benefits from more of them.
At test time, we encode the query once per photograph and average the resulting embeddings, without retraining.
For each number of photographs $k$ from 1 to 5, we evaluate every choice of $k$ photographs among the five (\eg, 10 choices for $k=2$) and report the mean mAP (\cref{fig:refcount}); for $k=1$, this mean over the five photographs differs slightly from the single default photograph of \cref{tab:ablation}.
More photographs consistently help: mAP rises from 43.6 to 45.0 for CAST-2B (+1.5) and from 50.1 to 52.0 for CAST-8B (+1.9), on single-person (+1.3 and +2.0) and multi-person queries (+2.1 and +1.6).
All gains from one to five photographs are significant ($p<0.002$, paired test over queries).
Most of the gain comes from the second photograph; beyond it, gains are smaller and not significant for CAST-8B on multi-person queries ($p=0.13$).
Averaging also makes CAST less sensitive to which photographs are chosen, as the spread across choices shrinks.
A single photograph thus gives most of the accuracy, and additional photographs bring a small gain at no training cost.

\section{Evaluation Details}
\label{sec:supp_eval}

\paragraph{Metrics.}
R@1 is a success rate: the fraction of queries whose top-ranked shot is a positive. R@5 and R@10 are recall over the positives: the fraction of a query's positives retrieved in the top 5 and top 10. On ITCPR, R@$k$ follows the protocol of FAFA~\cite{fafa}.

\paragraph{Seen and unseen actors.}
A test query belongs to the \emph{seen} group if all its referenced actors appear in training movies (135 queries), to the \emph{unseen} group if none does (610 queries); the 38 multi-person queries that mix both are excluded from this comparison.
These mixed queries score higher than both groups (\cref{tab:seen}), which is why the mean over all queries exceeds both group means.

\begin{table}[h]
  \centering
  \small
  \begin{tabular}{l cccc}
    \toprule
    Method & Seen (135) & Unseen (610) & Mixed (38) & All (783) \\
    \midrule
    Late Fusion & 33.8 & 36.1 & 50.6 & 36.4 \\
    CAST-2B & 42.9 & 43.2 & 46.6 & 43.3 \\
    CAST-8B & 45.4 & 50.0 & 59.6 & 49.7 \\
    \bottomrule
  \end{tabular}
  \caption{\textbf{mAP on queries with seen, unseen, and mixed actors.}}
  \label{tab:seen}
\end{table}

\paragraph{Training groups of the teacher.}
Each training group of the reranker draws three identity and three action negatives. When a query has fewer negatives of one type, the missing ones are drawn from the other type; when it has fewer than six in total, the group is smaller.

\end{document}